%% file: main.tex
\documentclass[10pt,twocolumn,letterpaper]{article}

\usepackage[pagenumbers]{cvpr} 
\usepackage[accsupp]{axessibility} 
\input{preamble}

\usepackage{enumitem}
\usepackage{bm}
\usepackage{url}
\usepackage{comment}
\usepackage{graphicx}
\usepackage{colortbl}
\usepackage{amsmath}
\usepackage{multirow}
\usepackage{adjustbox}
\usepackage{threeparttable}
\usepackage{epsfig}
\usepackage{wrapfig}
\usepackage{marvosym}
\usepackage{ifsym}
\definecolor{cvprblue}{rgb}{0.21,0.49,0.74}
\usepackage[pagebackref,breaklinks,colorlinks,citecolor=cvprblue]{hyperref}
\def\paperID{187} 
\def\confName{CVPR}
\def\confYear{2024}
\title{Video Super-Resolution Transformer with Masked Inter\&Intra-Frame Attention}

\author{\hspace{-0.5cm}
Xingyu Zhou$^1$ Leheng Zhang$^1$
Xiaorui Zhao$^1$  Keze Wang$^2$ Leida Li$^3$ Shuhang Gu$^1$\thanks{Corresponding author.}\\
\hspace{-0.5cm}
$^1$University of Electronic Science and Technology of China \hspace{0pt}
$^2$Sun Yat-sen University \hspace{0pt}
$^3$ Xidian University\\
\hspace{-0.5cm}
{\tt\small \{xy.chous526, lehengzhang12, zzzhaoxiaorui, kezewang, shuhanggu\}@gmail.com, ldli@xidian.edu.cn}}

\begin{document}
\maketitle
\begin{abstract}
Recently, Vision Transformer has achieved great success in recovering missing details in low-resolution sequences, i.e., the video super-resolution (VSR) task.
Despite its superiority in VSR accuracy, the heavy computational burden as well as the large memory footprint hinder the deployment of Transformer-based VSR models on constrained devices.
In this paper, we address the above issue by proposing a novel feature-level masked processing framework: VSR with \textbf{M}asked \textbf{I}ntra and inter-frame \textbf{A}ttention (MIA-VSR).
The core of MIA-VSR is leveraging feature-level temporal continuity between adjacent frames to reduce redundant computations and make more rational use of previously enhanced SR features.
Concretely, we propose an intra-frame and inter-frame attention block which takes the respective roles of past features and input
features into consideration and only exploits previously enhanced features to provide supplementary information.
In addition, an adaptive block-wise mask prediction module is developed to skip unimportant computations according to feature similarity between adjacent frames.
We conduct detailed ablation studies to validate our contributions and compare the proposed method with recent state-of-the-art VSR approaches.
The experimental results demonstrate that MIA-VSR improves
the memory and computation efficiency over state-of-the-art methods, without trading off PSNR accuracy. The code is available at \url{https://github.com/LabShuHangGU/MIA-VSR}.
\end{abstract} 
\section{Introduction}
\label{sec:intro}
Image super-resolution (SR) refers to the process of recovering sharp details in high resolution (HR) images from low resolution (LR) observations.
Due to its great value in practical usages, e.g., surveillance and high deﬁnition display, SR has been a thriving research topic over the past twenty years.
Generally, compared with single image super-resolution which only exploits intra-frame information to estimate the missing details, video super-resolution (VSR) additionally leverages the temporal information to recover the HR frames and therefore often leads to better SR results.

\begin{figure}
\hspace{-0.46cm}
\includegraphics[width=0.50\textwidth]{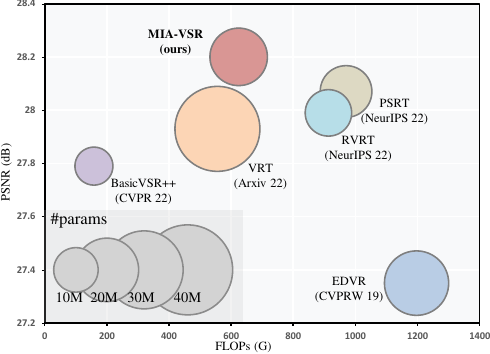}
\caption{\textbf{PSNR(dB) and FLOPs(G) comparison on the Vid4 \cite{liu2013bayesian} dataset.} We compare our MIA-VSR model with the state-of-the-art temporal sliding-window and recurrent based VSR models, including EDVR \cite{wang2019edvr}, BasicVSR++ \cite{chan2022basicvsr++}, VRT \cite{liang2022vrt}, RVRT \cite{liang2022recurrent} and PSRT \cite{shi2022rethinking}. Our MIA-VSR model outperforms these methods and strikes a balance between performance and compute efficiency.}
\label{fig:IIABwithmask}
\end{figure}
The key of VSR lies in making rational use of temporal information. 
Researchers in the early stages utilized convolutional neural networks (CNNs) to extract features and have investigated advanced information propagation strategies \cite{fuoli2019efficient,isobe2020video,chan2021basicvsr,chan2022basicvsr++,shi2022rethinking}, sophisticated alignment modules \cite{wang2019edvr,chan2021basicvsr,tian2020tdan,liang2022recurrent,shi2022rethinking,xu2023implicit}, effective training strategies \cite{xiao2021space, lin2023accelerating} as well as elaborately designed network architectures \cite{li2020mucan,isobe2022look} for the pursuit of highly accurate VSR results.
Recently, with the rapid development of Transformers in computer vision,  
several attempts have been made to exploit Transformers for better recovering missing details in LR sequences \cite{liang2022vrt,qiu2022learning,liu2022learning,liang2022recurrent, shi2022rethinking}. 
Due to the powerful representation learning capabilities of self-attention structure, these Transformer-based approaches have raised the state-of-the-art in VSR to a new level.

In spite of its superior SR results, the heavy computational burden and large memory footprint \cite{liang2022vrt, liang2022recurrent, shi2022rethinking} of Transformer-based VSR approaches limit their application to constrained devices.
With the common availability of video cameras, efficient video processing has become an increasingly important research topic in the literature on computer vision.
Paralleling to network pruning and quantization, which is applicable to all kinds of application, exploiting temporal redundancy for avoiding unnecessary computation is a specific strategy for accelerating video processing.
In their inspiring work, Habibian \etal\cite{habibian2021skip} proposed a skip-convolution method that restricts the computation of video processing only to regions with significant changes while skipping the others. 
Although the skip-covolution strategy could save computations without significant performance drop in high-level tasks, e.g. object detection and
human pose estimation, as low-level vision tasks such as video super-resolution are highly sensitive to minor changes in image content, 
whether such a skip-processing mechanism is applicable to VSR is still an open question.

In this paper, we provide an affirmative answer to the above question with a novel masked VSR framework, i.e., \textbf{M}asked \textbf{I}nter\&Intra frame \textbf{A}ttention (MIA) model.
Our MIA-VSR method could make use of temporal continuity to recover missing HR details, while, at the same time, leveraging temporal redundancy for reducing unnecessary  computations.
Concretely, our MIA-VSR model advances the existing VSR approaches in the following two aspects.
First, we develop a tailored inter-frame and intra-frame attention block (IIAB) for making more rational use of previously enhanced features.
Instead of directly inheriting the Swin-Transformer block \cite{liu2021swin} to process concatenated hidden states and image feature, our proposed IIAB block takes into account the respective roles of past features and input features and only utilizes the image feature of the current frame to generate the query token.
As a result, the proposed IIAB block not only reduces the computational consumption of the original joint self-attention strategy by a large margin, but also aligns the intermediate feature with spatial coordinates to enable efficient masked processing.
Secondly, we propose a feature-level adaptive masked processing mechanism
to reduce redundant computations according to the continuity between adjacent frames. 
Differently from previous efficient processing models, which simply determine skipable regions of the whole network according to pixel intensities, we adopt a feature-level selective skipping strategy and pass by computations of a specific stage adaptively.
The proposed feature-level adaptive masking strategy enables our MIA-VSR to save computations and at the same time achieve good VSR results.
Our contributions can be summarized as follows.

\begin{itemize}
\item We present a novel feature-level masked processing framework for efficient VSR, which is able to take advantage of the temporal continuity to reduce redundant computations in the VSR task.
\item We propose an intra-frame and inter-frame attention block, which could effectively extract spatial and temporal supplementary information to enhance SR features.
\item We propose an adaptive mask prediction module, which masks out unimportant regions according to the feature similarity between adjacent frames for different stages of processing in the VSR model.
\item We compare our proposed MIA-VSR model with state-of-the-art VSR models against which our approach could generate superior results with less computations and memory footprints.
\end{itemize}
\vspace{-0.2cm}
\section{Related Work}
\label{gen_inst}
\subsection{Video Super-Resolution}
According to how the temporal information is utilized, existing deep learning based VSR methods can be grouped into two categories: temporal sliding-window based methods and recurrent based methods.

\vspace{\baselineskip}
\noindent\textbf{Temporal sliding-window based VSR.}
Given an LR video sequence, a category of approaches processes the LR frames in a temporal sliding-window manner which aligns adjacent LR frames with the reference frame to estimate a single HR output.
The alignment module plays an essential role in temporal sliding-window based method.
Earlier work \cite{caballero2017real, liu2017robust, tao2017detail} explicitly estimated the optical flow to align adjacent frames.
Recently, implicit alignment modules \cite{xu2023implicit} were proposed to perform alignment in the high-dimensional feature space.
Dynamic filters \cite{jo2018deep}, deformable convolutions \cite{dai2017deformable, tian2020tdan, wang2019edvr} and attention modules \cite{Isobe_2020_CVPR, li2020mucan} have been developed to conduct motion compensation implicitly in the feature space.
In addition to the alignment module, another important research direction is investigating sophisticated network architectures to process aligned images.
Many works were aimed at estimating HR images from multiple input images in a temporal sliding-window manner \cite{li2020mucan, wang2019edvr, cao2021video, liang2022vrt}. 
%
%
Although the alignment module enables sliding window based VSR networks to better leverage temporal information from adjacent frames, accessible information to the VSR models is limited by the size of temporal sliding window and these methods could only make use of temporal information from limited number of input video frames.

\begin{figure*}
\centering
\includegraphics[width=0.96\textwidth]{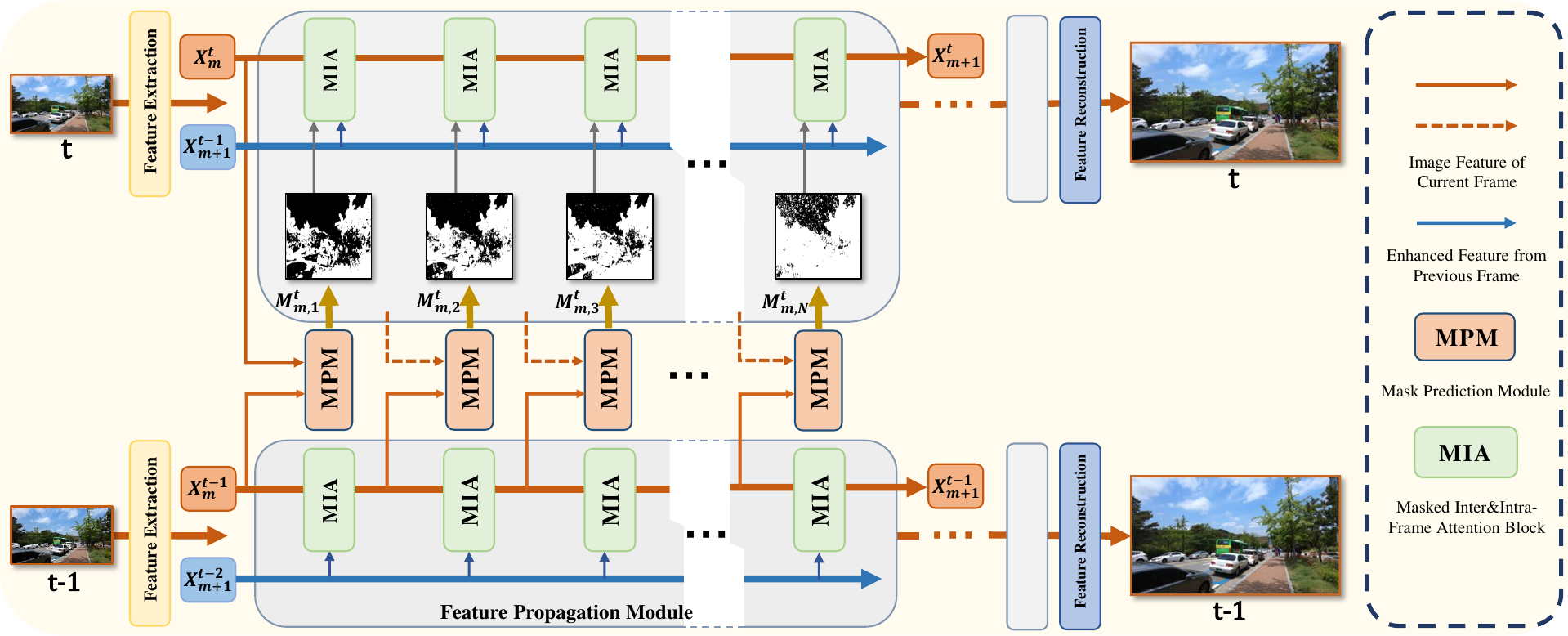}
\caption{\textbf{The overall architecture of MIA-VSR.} We develop a feature-level masked processing framework which uses the mask prediction module (MPM) to reduce redundant computations by leveraging temporal continuity, and propose a masked intra-frame and inter-frame (MIA) block to make more rational use of previous enhanced features to support the feature enhancement of the current frame. Our MIA-VSR model can be easily extended to the bi-directional second-order grid propagation framework as \cite{chan2022basicvsr++}. More details of our proposed MIA-VSR can be found in Section \ref{sec:Method}.}
\label{fig1}
\end{figure*}
\vspace{\baselineskip}
\noindent\textbf{Recurrent framework based VSR.}
Another category of approaches applies recurrent neural networks to exploit temporal information from more frames.
FRVSR~\cite{sajjadi2018frame} first proposed a recurrent framework that utilizes optical flow to align the previous HR estimation and the current LR input for VSR.
%
%
RLSP~\cite{fuoli2019efficient} propagates high-dimensional hidden states instead of the previous HR estimation to better exploit long-term information.
RSDN~\cite{isobe2020video} further extended RLSP~\cite{fuoli2019efficient} by decomposing the LR frames into structure and detail layers and introduced an adaptation module to selectively use the information from hidden state.
BasicVSR \cite{chan2020basicvsr} utilized bi-directional hidden states, and BasicVSR++ \cite{chan2022basicvsr++} further improved BasicVSR with second-order grid propagation and flow-guided deformable alignment.
PSRT \cite{shi2022rethinking} adopted the bi-directional second-order grid propagation framework of BasicVSR++ and utilized multi-frame self-attention block to jointly process the outputs of the feature propagation of the previous frames and the input features.
Generally, by passing the high dimensional hidden states or output feature, the recurrent based VSR could incorporate temporal information from more frames for estimating the missing details and therefore achieve better VSR results.
Our proposed MIA-VSR follows the general framework of bi-directional second-order hidden feature propagation while introducing masked processing strategy and intra\&inter-frame attention block for the pursuit of better trade-off between VSR accuracy, computational burden and memory footprint.
\subsection{Efficient Video Processing}
Various strategies of reducing temporal redundancy have been explored for efficient video processing.
A category of approaches adopts the general network optimization strategy and utilizes pruning \cite{xia2022residual} or distillation \cite{habibian2022delta} methods to train light-weight networks for efficient video processing.
In order to take extra benefit from the temporal continuity of video frames, another category of methods only computes expensive backbone features on key-frames and applies concatenation methods \cite{jain2019accel}, optical-flow based alignment methods \cite{zhu2017deep,li2018low,zhu2018towards,jain2019accel,nie2019dynamic,hu2020temporally}, dynamic convolution methods \cite{nie2019dynamic} and self-attention methods \cite{hu2020temporally,liu2022learning} to enhance features of other frames with keyframe features.
Most recently, \cite{habibian2021skip} proposed a skip-convolution approach which only conducts computation in regions with significant changes between frames to achieve the goal of efficient video processing.
However, most of the advanced efficient video processing schemes described above were designed for high-level vision tasks such as object detection and pose estimation. 
To the best of our knowledge, our study is the first work that leverages the temporal continuity across different areas for each block to reduce redundant computation for low-level VSR tasks.
\section{Methodology}
\label{sec:Method}

\begin{figure*}
\centering
\includegraphics[width=0.94\textwidth,height=0.37\textwidth]{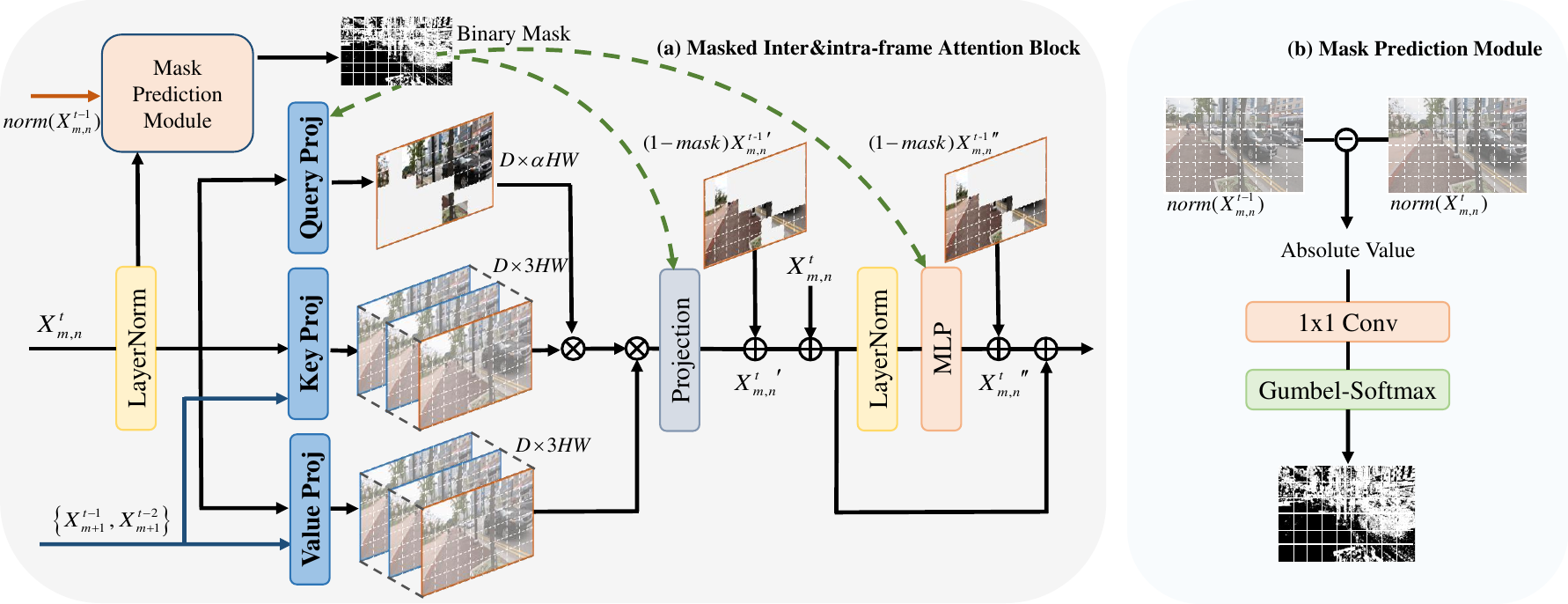}
\caption{\textbf{Illustration of the inter\&intra-frame attention block with adaptive masked processing module.} The adaptive mask prediction module \textbf{(b)} in the IIAB block acts in the \texttt{Attention} module's linear layer which produce the Query, the projection layer and the linear layer in the \texttt{FFN} module during inference to reduce temporal and sptical redundancy calculations \textbf{(a)}. ${\bm{X}_{m,n}^t}^{\prime}$ and ${\bm{X}_{m,n}^t}^{\prime\prime}$ refer to the processed hidden feature in the \texttt{Attention} and \texttt{FFN} module.}
\label{IIABwithmask}
\end{figure*}
\subsection{Overall Architecture}
Given $T$ frames of low-resolution video sequence $\bm{I}_{LR} \in  \mathbb{R}^{T \times H \times W \times 3}$, our goal is to reconstruct the corresponding HR video sequence
$\bm{I}_{HR}\in \mathbb{R}^{T \times sH \times sW \times 3}$, where $s$ is the scaling factor and $H$, $W$, $3$ are the height, width and channel number of input frames.

The overall architecture of our proposed MIA-VSR model is shown in Fig.\ref{fig1}. We built our MIA-VSR framework upon the bi-directional second-order grid propagation framework of BasicVSR++ \cite{chan2022basicvsr++}, which has also been adopted in the recent state-of-the-art method PSRT \cite{shi2022rethinking}.
The whole model consists of three parts, i.e., the shallow feature extraction part, the recurrent feature refinement part and the feature reconstruction part.
We follow previous works BasicVSR++ \cite{chan2022basicvsr++} and PSRT \cite{shi2022rethinking} which use a plain convolution operation to extract shallow features and adopt a pixel-shuffle layer \cite{shi2016real, gu2019self} to reconstruct HR output with refined features.
The patch alignment method used in PSRT \cite{shi2016real} 
is used to align the adjacent frames.
We improve the efficiency of existing works by reducing redundant computations in the recurrent feature refinement part.

Generally, the recurrent feature refinement part comprises $\textbf{M}$  feature propagation modules and each feature propagation module consists of $ \textbf{N}$ cascaded processing blocks.
The feature propagation module takes the enhanced outputs from previous frames as well as image feature of the current frame as input.
For the $t$-th frame, let us denote the input feature for the $m$-th feature propagation module as $\bm{X}_m^t$.
The feature propagation module takes $\bm{X}_m^t$, $\bm{X}_{m+1}^{t-1}$ and $\bm{X}_{m+1}^{t-2}$ as inputs
to calculate the enhanced feature:
\begin{equation}\label{eq:mask}
    \bm{X}_{m+1}^t = \texttt{FPM}(\bm{X}_m^t, \bm{X}_{m+1}^{t-1}, \bm{X}_{m+1}^{t-2}),
\end{equation}
where $\bm{X}_{m+1}^{t}$ is the enhanced output of the current frame, 
$\bm{X}_{m+1}^{t-1}$ and $\bm{X}_{m+1}^{t-2}$ are enhanced feature from the past two frames.
To be more specific, in each feature propagation module, cascaded processing blocks utilize outputs from the previous frames to enhance input features.
In comparison to the previous methods that directly inherit the Swin transformer block to process concatenated features, we propose a tailored  intra\&inter frame attention block (IIAB)  
to more efficiently enhance $\bm{X}_m^t$ with the help of $\bm{X}_{m+1}^{t-1}$ and $\bm{X}_{m+1}^{t-2}$:
\begin{equation}\label{eq:mask}
    \bm{X}_{m,n+1}^t = \texttt{IIAB}(\bm{X}_{m,n}^t, \bm{X}_{m+1}^{t-1}, \bm{X}_{m+1}^{t-2}),
\end{equation}
where $\bm{X}_{m,n}^t$ is the input to the $n$-th IIAB in the $m$-th feature propagation module, $\bm{X}_{m,0}^t=\bm{X}_{m}^t$ and $\bm{X}_{m,N}^t=\bm{X}_{m+1}^t$ are the input and output of the whole FPM, respectively. 
More details of the proposed IIAB will be introduced in the following subsection \ref{sec:iiab}.

In order to reduce redundant computations according to temporal continuity,
we further develop an adaptive mask prediction module to generate block-wise masks $\bm{M}_{m,n}^t$,
with which 
we could directly utilize the outputs from past frame and selectively skip unimportant computation:
\begin{equation}
\label{eq:maskedprocessing}
    \bm{\hat{X}}^t_{m,n} = \bm{M}_{m,n}^{t}\odot \bm{X}^{t}_{m,n} + (1-\bm{M}_{m,n}^{t})\odot \bm{\hat{X}}^{t-1}_{m,n},
\end{equation}
where $\bm{\hat{X}}^t_{m,n}$ is the processed hidden feature in the IIAB block (shown as ${\bm{X}_{m,n}^t}^{\prime}$ and ${\bm{X}_{m,n}^t}^{\prime\prime}$ in Fig.\ref{IIABwithmask}) and $\odot$ is the point-wise multiplication operation.
More details of our mask predicting module will be presented in Section \ref{sec:mask}.

\subsection{Inter\&Intra-Frame Attention Block}
\label{sec:iiab}
As introduced in the previous section, cascade processing blocks play a key role in extracting supplementary information from previous frames to enhance SR features.
To achieve this goal, the previous method \cite{liang2022recurrent,shi2022rethinking} simply adopts the multi-head self-attention block with shifted local windows in the Swin Transformer to process concatenated hidden states (enhanced features of previous frames) and the current input feature. 
In this paper, we take the respective role of the previously enhanced feature and the current feature into consideration and 
propose a intra\&inter frame attention block (IIAB) for efficient VSR.
To be more specific, we think the enhanced features of the previous frames $\bm{X}_{m+1}^{t-1}$ and $ \bm{X}_{m+1}^{t-2}$ should only used for providing supplementary information and do not need to be further enhanced.
Therefore, we only utilize feature of the current frame to generate Query Tokens, and adopt enhanced features from the previous frames as well as feature of the current frame to generate Key and Value Tokens:
\begin{equation}
\label{eq:QKV}
    \begin{aligned}
    &\bm{Q}_{m,n}^{t} = \bm{X}_{m,n}^{t}\bm{W}^Q_{m,n},\\
    &\left\{
        \begin{array}{l}
        \bm{K}_{m,n}^{t,intra} =  \bm{X}_{m,n}^{t}\bm{W}^K_{m,n},\\
        \bm{K}_{m,n}^{t,inter} = \left[\bm{X}_{m+1}^{t-1}; \bm{X}_{m+1}^{t-2}\right]\bm{W}^K_{m,n},
        \end{array}
    \right.\\
    &\left\{
        \begin{array}{l}
        \bm{V}_{m,n}^{t, intra} =  \bm{X}_{m,n}^{t}\bm{W}^V_{m,n},\\
        \bm{V}_{m,n}^{t, inter} = \left[\bm{X}_{m+1}^{t-1}; \bm{X}_{m+1}^{t-2}\right]\bm{W}^V_{m,n},
        \end{array}
     \right.\\
     \end{aligned}
\end{equation}
where $\bm{W}^Q_{m,n}$, $\bm{W}^K_{m,n}$ and $\bm{W}^V_{m,n}$ are the respective projection matrices; 
$\bm{Q}_{m,n}^{t}$ represents the Query Tokens generated from the current input feature;
 $\bm{K}_{m,n}^{t, intra}$ and $\bm{K}_{m,n}^{t, inter}$ are the intra-frame and inter-frame Key Tokens and $\bm{V}_{m,n}^{t, intra}$ and $\bm{V}_{m,n}^{t, inter}$ are the intra-frame and inter-frame Value Tokens.
IIAB jointly calculates the attention map between query token and intra\&inter-frame keys to generate the updated feature:
\begin{equation}
\label{eq:SA}
\setlength\abovedisplayskip{1pt}
\setlength\belowdisplayskip{1pt}
      \texttt{IIAB}_\texttt{Attention} = \texttt{SoftMax}(\bm{Q}_{m,n}^{t}{\bm{K}_{m,n}^{t}}^T/\sqrt{d}+B)\bm{V}_{m,n}^{t},
\end{equation}
where $\bm{K}_{m,n}^{t} = [\bm{K}_{m,n}^{t,inter};\bm{K}_{m,n}^{t,intra}]$ and $\bm{V}_{m,n}^{t} = [\bm{V}_{m,n}^{t,inter};\bm{V}_{m,n}^{t,intra}]$ are the concatenated intra\&inter-frame tokens; $d$ is the channel dimension of the Token and $B$ is the learnable relative positional encoding.
It should be noted that in Eq.\ref{eq:QKV}, all the intermediate inter-frame tokes $\{\bm{V}_{m,n}^{t,inter};\bm{K}_{m,n}^{t,inter}\}_{n=1,\dots, N}$ are generated from the same enhanced features $\left[\bm{X}_{m+1}^{t-1}; \bm{X}_{m+1}^{t-2}\right]$;
which means that we only leverage mature enhanced features from previous frames to provide supplementary information and do not need to utilize the time consuming self-attention mechanism to jointly update the current feature and previous features.
An illustration of our proposed inter-frame and intra-frame attention (IIAB) block is presented in Fig.\ref{IIABwithmask}.
In addition to the attention block, our proposed transformer layer also utilizes LayerNorm and FFN layers, which have been commonly utilized in other Transformer-based VSR architectures \cite{liang2022recurrent, shi2022rethinking}.

In our implementation, we also adopt the Shift-window strategy to process the input LR frames and the above self-attention calculation is conduct in $W\times W$ non-overlapping windows.
Denote our channel number by $C$, $\bm{Q}_{m,n}^{t}$, $\bm{K}_{m,n}^{t}$ and $\bm{V}_{m,n}^{t}$ in Eq.\ref{eq:SA} are with sizes of $N^2\times C$, $3N^2\times C$ and $3N^2\times C$, respectively. 
Therefore, Eq.\ref{eq:SA} only suffers the calculation $1/3$ in the self-attention calculation procedure in comparison with the joint self-attention processing strategy.
Moreover, as we will validate in the ablation study section, our strategy not only avoids unnecessary computation in jointly updating previous features, but also provides better features for the final VSR task.

\subsection{Adaptive Masked Processing in the Recurrent VSR Transformer}
\label{sec:mask}
In this subsection, we present details of how we generate block-wise masks $\bm{M}_{m,n}^{t}$ in Eq.\ref{eq:maskedprocessing} to further reduce unimportant computations.
Habibian \etal\cite{habibian2021skip} first proposed a skip convolution mechanism to reduce redundant computations based on differences between pixel values in adjacent frames.
However, for the VSR task, skipping the whole computation process of a certain region according to the intensity differences between adjacent input frames will
inevitably affect the accuracy of the model. 
In this paper, we leverage the temporal continuity to reduce the computation of the VSR model in a subtler way: by exploiting feature differences between adjacent frames to select less important computations for each block.

Since features of different stages have various contributions to the final VSR result, using
feature differences between adjacent frames to generate binary masks through a uniform threshold is non-trivial.
Therefore, we propose an adaptive masked processing scheme which jointly trains tiny mask predicting networks with VSR feature enhancement blocks.
To be more specific, in each stage of processing, the mask predicting network takes  the difference between normalized features as input:
\begin{equation}\label{eq:mask}
    \Delta \bm{X}_{m,n}^{t-1\to t} = \|\texttt{norm}(\bm{X}^t_{m,n}) - \texttt{norm}(\bm{X}^{t-1}_{m,n})\|_1.
\end{equation}
Then, we employ  a Gumbel-softmax \cite{jang2016categorical} gate to sample masking features:
\begin{equation}
\left\{
    \begin{aligned}
    &\texttt{Mask}(\Delta \bm{X}_{m,n}^{t-1\to t}))=\frac{\exp \left(\left(\log \left(\pi_1\right)+g_1\right) / \tau\right)}{\sum_{i=1}^{2} \exp \left(\left(\log \left(\pi_i\right)+g_{i}\right) / \tau\right)},\\
    &\pi_1 = \texttt{Sigmoid}(f(\Delta \bm{X}_{m,n}^{t-1\to t})), \\
    &\pi_2 = 1-\texttt{Sigmoid}(f(\Delta \bm{X}_{m,n}^{t-1\to t})),
    \end{aligned}
    \right.
\end{equation}
where $f(\cdot)$ is a $1\times1$ convolution layer to weighted sum feature differences from different channels;
$\pi_1$ could be interpreted as the probability of whether a position should be preserved;
and  $g_1$, $g_2$ are noise samples drawn from a Gumbel distribution, $\tau$ 
 is the temperature coefficient and we set it as 2/3 in all of our experiments.
 With the masking feature $\texttt{Mask}(\Delta \bm{X}_{m,n}^{t-1\to t}))$, we could directly set a threshold value to generate the binary mask:
 \begin{equation}\label{eq:mask}
    \bm{M}_{m,n}^t = 
    \left\{
        \begin{array}{l}
        1,\quad\quad if \quad \texttt{Mask}(\Delta \bm{X}_{m,n}^{t-1\to t}) > 0.5; \\
        0,\quad\quad else.
        \end{array}
        \right.
\end{equation}
The above Gumbel-softmax trick enables us to train mask predicting networks jointly with the 
VSR network, 
for learning diverse masking criterion for different layers from the training data.
In the inference phase, we do not add Gumbel noise to generate the masking feature and directly generate binary masks with $f(\Delta \bm{X}_{m,n}^{t-1\to t})$.

We apply the same mask to save computations in the $\texttt{IIAB}_\texttt{Attention}$ part and the following $\texttt{FFNs}$ in the feature projection part.
The above masked processing strategy allows us to generate less Query tokens in the $\texttt{IIAB}_\texttt{Attention}$ part and skip projections in the feature projection part.
Let's denote $\alpha\in[0,1]$ as the percentage of
non-zeros in $\bm{M}_{m,n}^t$, our Masked Intra\&Inter-frame Attention Block is able to reduce the computational complexity of each IIAB from $12HWC^2+(4+12N^2)HWC$ to $(6+6\alpha)HWC^2+(3\alpha N^2+9N^2+4)HWC$.
The saved computations overwhelm the extra computations introduced by the tiny mask predicting networks, which only use one layer of $1 \times 1$ convolution to reduce the channel number of the feature difference tensor from $C$ to 1. 
\subsection{Training Objectives}
We train our network in a supervised manner.
Following recent state-of-the-art approaches, we utilize the
Charbonnier loss \cite{charbonnier1994two} $\mathcal{L}_{sr} = \sqrt{\parallel\bm{\hat{I}}^{HQ} -\bm{I}^{HQ} \parallel^2 + \varepsilon ^2}$ between the estimated HR image $\bm{\hat{I}}^{HQ}$ and the ground truth image $\bm{I}^{HQ}$ to train our network; 
where $\epsilon$ is a constant and we set it as $10^{-3}$ in all of our experiments.
Moreover, In order to push our mask predicting networks to mask out more positions, we apply a $\ell_1$ loss on the masking features:
\begin{equation}
     \mathcal{L}_{mask} = \frac{1}{MNZ}\sum\nolimits_{m = 1}^{M}\sum\nolimits_{n = 1}^{N}\|\texttt{Mask}(\Delta \bm{X}_{m,n}^{t-1\to t}))\|_1,
\end{equation}
where $Z$ is the number of pixels for each masking feature.
Our network is trained by a combination of $\mathcal{L}_{sr}$ and $\mathcal{L}_{mask}$:
\begin{equation}\label{eq:loss}
     \mathcal{L} = \mathcal{L}_{sr}+\lambda\mathcal{L}_{mask},
\end{equation}
 $\lambda$ is a factor for masking ratio adjustment.
More details can be found in our experimental section.
\section{Experiments}
\subsection{Experimental Settings}
Following the experimental settings of recently proposed VSR methods \cite{shi2022rethinking,liang2022recurrent,liang2022vrt,chan2022basicvsr++,cao2021video}, 
we evaluate our MIA-VSR model on the REDS \cite{nah2019ntire}, Vimeo90K \cite{xue2019video} and the commonly used Vid4 \cite{liu2013bayesian} datasets.
We train two models on the REDS dateset and the Vimeo90K dataset, respectively.
The model trained on the REDS dataset is used for evaluating the REDS testing data and the Viemo90K model is used for evaluating the Vimeo90K and Vid4 testing data.
For our proposed adaptive mask prediction module, we fine-tune it with the pre-trained VSR model for another 100K iterations. 
We implement our model with Pytorch and train our models with RTX 4090 GPUs.
The respective hyper-parameters used for ablation study and comparison with state-of-the-art methods will be introduced in the following subsections.

\subsection{Ablation Study}
\label{sec:ablation}
\paragraph{The effectiveness of IIAB}
\begin{table}
	\centering
 \caption{Ablation studies on the processing blocks and the proposed adaptive mask prediction module. More details can be found in our ablation study section.} 
	\label{tab:threshold_ablation}
    \begin{threeparttable}
    \begin{adjustbox}{center}  
    \centering
    \scalebox{0.82}{\begin{tabular}{c|c|c||ccc}
    \toprule
    \multirow{2}{*}{Model} &\multirow{2}{*}{ $\lambda$}&Params& \multicolumn{3}{c}{REDS4} \\
    &&(M)& PSNR & SSIM &FLOPs(G)    \\
    \midrule
    MFSAB-VSR & - & 6.41 & 31.03 & 0.8965 & 871.59 \\
    IIAB-VSR & - & 6.34 & 31.12 & 0.8979 & 518.92 \\
    \midrule
    \midrule
    IIAB-VSR + HM & - & 6.34& 29.83 & 0.8390 & 420.53 \\
    MIA-VSR & 1e-4& 6.35&31.11& 0.8978& 506.28\\
    MIA-VSR & 3e-4& 6.35&31.07&0.8972 &469.78  \\
    MIA-VSR & 5e-4& 6.35& 31.01 & 0.8966 & 442.32  \\
    MIA-VSR & 1e-3& 6.35& 30.76& 0.8773 & 426.84  \\
    \bottomrule
  \end{tabular}}
  \end{adjustbox}
  \end{threeparttable}
\end{table}
\begin{figure}
\hspace{-0.62cm}
\includegraphics[width=0.50\textwidth]{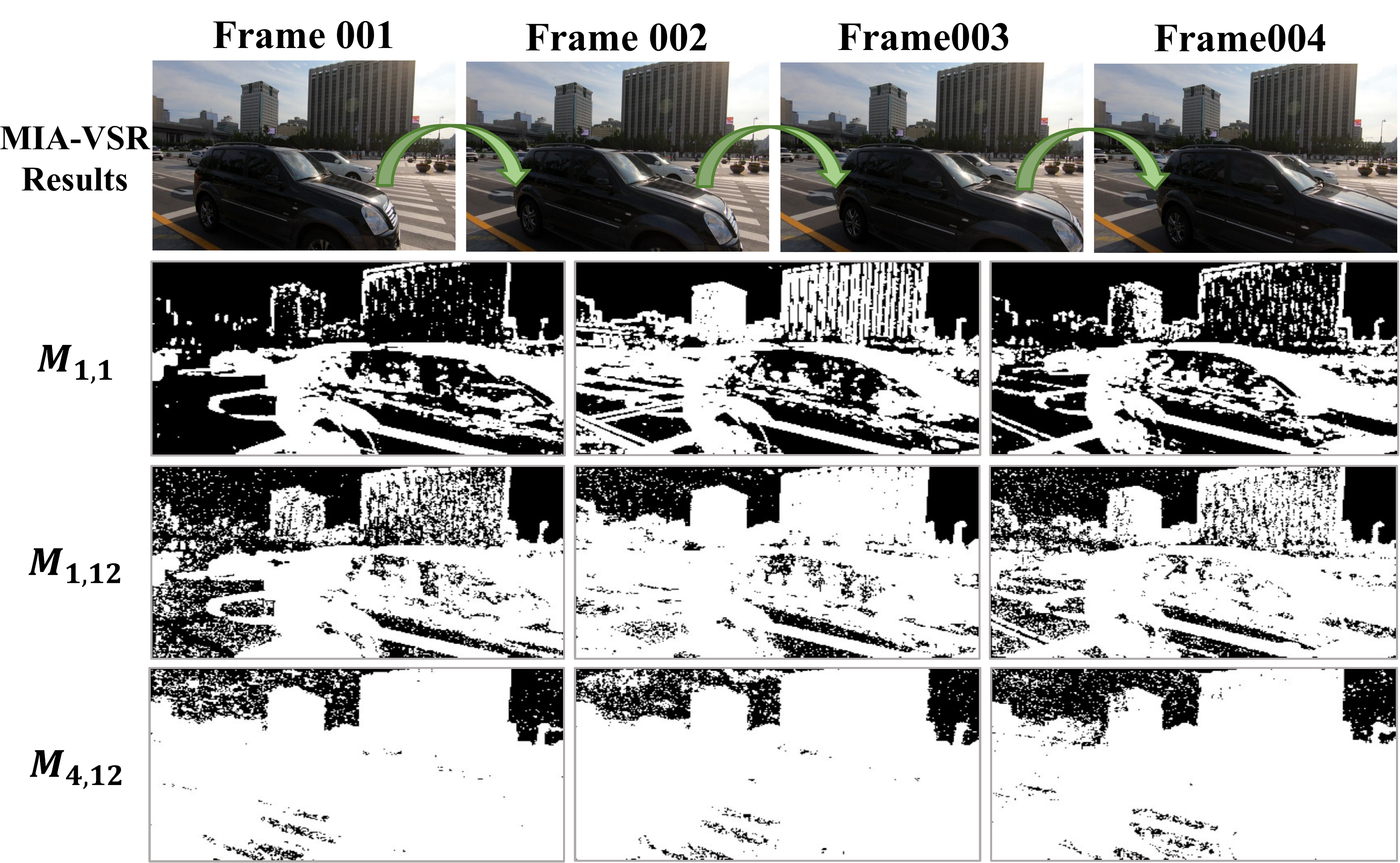}
\caption{Visualization of predicted masks for a sequence in the REDS dataset.}
\label{fig:IIABwithmask}
\end{figure}
In order to show the advantages of the proposed IIAB, we firstly compare the proposed intra\&inter-frame attention block (IIAB) with the multi-frame 
self-attention block (MFSAB) which was adopted in PSRT \cite{shi2022rethinking}.
We use 6 IIAB or MFSAB blocks to build feature propagation modules and instantialize VSR models with 4 feature propagation modules.
The channel number, window size and the head number for the two models are set as 120, 8 and 6, respectively.
We denote the two VSR models as IIA-VSR, MFSA-VSR, and train the two models with 6 frame training data for 300K iterations.
The super-resolution results of the two models are shown in Table\ref{results-table}.
The number of parameters and computational consumption (FLOPs) of the two models are also reported for reference.
The number of FLOPs is calculated on the REDS dataset, which super-resolve $180 \times 320$ video sequences to a resolution of $720 \times 1280$, we report the average FLOPs for each frame of processing.
\begin{table*}
	\renewcommand{\arraystretch}{0.82}
	\renewcommand{\tabcolsep}{4pt}
    \caption{Quantitative comparison (PSNR/SSIM) on the REDS4\cite{nah2019ntire}, Vimeo90K-T\cite{xue2019video} and Vid4\cite{liu2013bayesian} dataset for 4× video super-resolution task. For each group of experiments, we color the best and second best performance with \textcolor{red}{red} and \textcolor{blue}{blue}, respectively.}
    \label{tab:sota-table}
    \begin{adjustbox}{center}
    \centering
    \begin{tabular}{c|c||ccc|ccc|ccc}
    \toprule
    \multirow{2}{*}{Method} & Frames  & \multicolumn{3}{c}{REDS4} & \multicolumn{3}{c}{Vimeo-90K-T} & \multicolumn{3}{c}{Vid4}  \\
    & REDS/Vimeo& PSNR & SSIM&FLOPs & PSNR & SSIM&FLOPs & PSNR & SSIM&FLOPs    \\
    \midrule
    TOFlow\cite{xue2019video} & 5/7 & 27.98 & 0.7990&- &33.08 & 0.9054 &-&25.89&0.7651& - \\
    EDVR\cite{wang2019edvr} & 5/7 & 31.09 & 0.8800&2.95 & 37.61 & 0.9489 &0.367& 27.35& 0.8264&1.197\\
    MuCAN\cite{li2020mucan} & 5/7 & 30.88 & 0.8750& 1.07& 37.32 & 0.9465 &0.135& -&-&- \\
    VSR-T\cite{cao2021video} & 5/7 &31.19 & 0.8815& -& 37.71 & 0.9494&- & 27.36&0.8258& -\\
    PSRT-sliding\cite{shi2022rethinking} & 5/- &  31.32 & 0.8834 & 1.66 & - & -& -& -& -& - \\
    VRT\cite{liang2022vrt} & 6/- &31.60 & 0.8888 & 1.37 & - & -& -& -& -& - \\
    PSRT-recurrent\cite{shi2022rethinking} & 6/- & 31.88 & 0.8964 & 2.39 & - & -&  -& -& -& -\\
    \rowcolor{gray!20}
    MIA-VSR(ours) & 6/- & \textcolor{red}{32.01} & \textcolor{red}{0.8997} & 1.50 & - & -& -& -& -& -  \\
    \midrule
    BasicVSR\cite{chan2020basicvsr}     & 15/14  & 31.42 & 0.8909& 0.33& 37.18 & 0.9450& 0.041 & 27.24 & 0.8251&0.134\\
    IconVSR\cite{chan2020basicvsr}      & 15/14  & 31.67 & 0.8948& 0.51& 37.47 & 0.9476& 0.063 & 27.39 & 0.8279&0.207\\
    TTVSR\cite{liu2022learning}&\textbf{50}/- &32.12&0.9021&0.61&-&-&-&-&-&-\\
    BasicVSR++\cite{chan2022basicvsr++} & \textbf{30}/14  & 32.39 & 0.9069& 0.39& 37.79 & 0.9500& 0.049 & 27.79 & 0.8400&0.158 \\
    VRT\cite{liang2022vrt}              & 16/7   & 32.19 & 0.9006& 1.37& 38.20 & 0.9530& 0.170 & 27.93 & 0.8425&0.556\\
    RVRT\cite{liang2022recurrent}       & \textbf{30}/14  & \textcolor{blue}{32.75} & \textcolor{blue}{0.9113}& 2.21& 38.15 & 0.9527& 0.275 & 27.99 & 0.8462&0.913 \\
    PSRT-recurrent\cite{shi2022rethinking} & 16/14  & 32.72 & 0.9106&2.39 &\textcolor{red}{38.27} & \textcolor{red}{0.9536}&0.297&\textcolor{blue}{28.07}& \textcolor{blue}{0.8485}&0.970\\
    \rowcolor{gray!20}
    MIA-VSR(ours)& 16/14  & \textcolor{red}{32.78} & \textcolor{red}{0.9220}&1.61 & \textcolor{blue}{38.22}& \textcolor{blue}{0.9532} &0.204& \textcolor{red}{28.20} & \textcolor{red}{0.8507}&0.624\\
    \bottomrule
    \end{tabular}
  \end{adjustbox}
\end{table*}
\begin{figure*}
\centering
\includegraphics[width=0.86\textwidth]{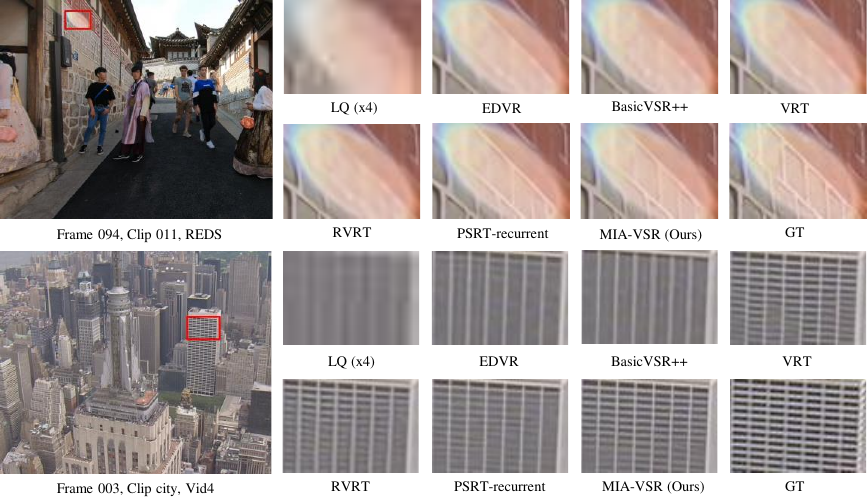}
\caption{Visual comparison for $4\times$ VSR on REDS4 dataset and Vid4 dataset.}
\label{fig:sotavisual}
\end{figure*}
As shown in Table \ref{tab:threshold_ablation}, it is clear that the proposed IIAB could generate better VSR results with less computational consumption than the MFSAB block. 
\paragraph{Masked processing strategy}
In this part, we validate the effectiveness of the proposed masked processing strategy.
Our basic experimental settings are the same as our experiment to validate IIAB.
We firstly present experimental results to test whether we could get good experimental results by setting a uniform threshold to generate masks based on feature differences.
We denote the model as IIAB-VSR + HM and set the threshold value as 0.2, where HM is the abbreviation for the handcrafted mask.

As can be found in Table \ref{tab:threshold_ablation}, adopting a uniform threshold for different layers will lead to a significant accuracy drop of the VSR model.
Then we validate the effectiveness of our proposed adaptive mask prediction module and analyze the effects of different $\lambda$ values in Eq.\ref{eq:loss}.
Concretely, we set $\lambda$ as $1e-4$, $3e-4$, $5e-4$ and $1e-3$ and train four different models.
Generally, setting a larger weight for the sparsity loss could push the network to mask our more computations but also results in less accurate VSR results.
By setting $\lambda$ as $5e-4$, we could further save approximately 20\% computations from the highly efficient IIAB-VSR model without a significant performance drop.
The much better trade-off between accuracy and computation by our MIA-VSR model over IIA-VSR + Handcraft Mask clearly validated the superiority of our adaptive mask predicting network.
Some visual examples of the masks generated in our MIA-VSR model can be found in Fig.\ref{fig:IIABwithmask}.
The network tends to skip a large portion of computations in the early stage and could mask fewer positions for deeper layers.
\subsection{Comparison with State-of-the-art  Methods}
In this subsection, we compare the proposed MIA-VSR model with state-of-the-art VSR methods.
We compare the MIA-VSR with representative sliding-window based methods TOFlow \cite{xue2019video}, EDVR \cite{wang2019edvr}, MuCAN \cite{li2020mucan}, VSR-T \cite{cao2021video}, VRT \cite{liang2022vrt}, RVRT \cite{liang2022recurrent} and representative recurrent-based methods BasicVSR \cite{chan2020basicvsr},  BasicVSR++ \cite{chan2022basicvsr++}, TTVSR \cite{liu2022learning}, PSRT-recurrent \cite{shi2022rethinking}; among which VRT, RVRT, TTVSR and PSRT-recurrent are Transformer-based approaches and the other approaches are CNN-based models. 

In order to achieve comparable VSR results with state-of-the-art methods, we instantialize our MIA-VSR model with 4 feature propagation modules and each feature propagation module contains 24 MIIA blocks. 
Among them, we set the interval of skip connections to [6,6,6,6].
The spatial window size, head size and channel size are set to $8\times 8$, 6 and 120 accordingly.
The number of parameters in our model is on par with the recent state-of-the-art methods RVRT \cite{liang2022recurrent} and PSRT-recurrent \cite{shi2022rethinking}.
Previous works \cite{liang2022vrt,liang2022recurrent, shi2022rethinking,liu2022learning,qiu2022learning} have figured out that models trained with longer sequences could achieve better VSR results.
For a fair comparison, we follow the experimental setting of \cite{liang2022recurrent,shi2022rethinking,liang2022vrt} and train the VSR model with short sequences (6 frames from the REDS dataset) and long sequences (16/14 frames from the REDS dataset and the Vimeo-90K dataset). 

The VSR results by different methods can be found in Table \ref{tab:sota-table}.
As the number of FLOPs(T) for our MIA-VSR model is related to the content of video sequences, we therefore report the average per frame FLOPs on different datasets.
In terms of the results trained with short sequences, the proposed MIA-VSR outperforms the compared methods by a large margin.
Our model improves the PSNR of the state-of-the-art PSRT-recurrent model by 0.13 dB with a reduction in the number of FLOPs of almost 40\%.
As for the models trained with longer training sequences, our MIA-VSR still achieves a better trade-off between VSR accuracy and efficiency over recent state-of-the-art approaches.
With more than 40\% less computations on the RVRT and PSRT-recurrent approaches, our model achieved the best VSR results
on the REDS and Vid4 datasets and the second-best results on the Vimeo-90K dataset.
Some visual results from different VSR results can be found in Fig.\ref{fig:sotavisual}, our proposed MIA-VSR method is able to recover more natural and sharp textures from the input LR video sequences.

In our supplementary file, we also provide other instantializations of our proposed MIA-VSR model to show the scalability of our approach and compare our method with efficient VSR models.
More visual examples are also provided for different datasets.

\subsection{Complexity and Memory Analysis}
In this part, we compare the complexity and memory footprint of different Transformer-based state-of-the-art VSR methods.
In Table \ref{tab:running time}, we report the number of parameters, the peak GPU memory consumption, the number of FLOPs, the Runtime and the PSNR by different VSR methods.
In comparison with other Transformer-based VSR methods, our MIA-VSR method has a similar number of parameters and requires much less number of FLOPs for processing the video sequence.
In addition, the peak GPU memory consumption, which is critical factor for deploying model on terminal equipment, by our model is much less than the VRT and RVRT approaches.
As for the runtime, our model is not as fast as RVRT, because the authors of RVRT have implemented the key components of RVRT with customized CUDA kernels.
As the acceleration and optimization of
Transformers are still to be studied, there is room for further optimization of the runtime of our method by our relatively small FLOPs.
\begin{table}
    \caption{Comparison of model size, testing memory and complexity of different VSR models on the  REDS\cite{nah2019ntire} dataset.}    \label{tab:running time}
    \renewcommand{\arraystretch}{1.2}
	\renewcommand{\tabcolsep}{2.5pt}
    \label{results-table}
    \begin{threeparttable}
    \begin{adjustbox}{center}  
    \centering
    \scalebox{0.76}{\begin{tabular}{c|c|c|c|c|c}
    \toprule
    Model&Params(M) &FLOPs(T)&Memory(M)&Runtime(ms)  &PSNR(dB)    \\
    \midrule
    VRT &35.6&1.37&2149& 888 &32.19    \\
    RVRT* &10.8&2.21&1056 & 473& 32.75    \\
    PSRT &13.4&2.39&190 & 1041&32.72 \\
    MIA-VSR &16.5&1.61 &206 & 822 & 32.78    \\
    \bottomrule
  \end{tabular}}
  \end{adjustbox}
  \begin{tablenotes} 
	\footnotesize \item * means that uses customized CUDA kernels for better performance.
    \end{tablenotes}
  \end{threeparttable}
\end{table}
\section{Conclusion}
In this paper, we proposed a novel Transformer-based recurrent
video super-resolution model, named MIA-VSR.
We proposed a masked processing framework to leverage the temporal continuity between adjacent frames to save computations for the video super-resolution model.
An Intra-frame and Inter-frame attention block is proposed to make better use of previously enhanced features to provide supplementary information; 
and an adaptive mask prediction module is developed to generate block-wise masks for each stage of processing.
Furthermore, we evaluated our MIA-VSR model on various benchmark datasets.
Our model is able to achieve state-of-the-art video super-resolution results with less computational resources.
\clearpage
{
    \small
    \bibliographystyle{ieeenat_fullname}
    \bibliography{main}
}
\maketitlesupplementary
\appendix
In this file, we provide more implementation and experimental details which are not included in the main text. 
In Section \ref{a}, we provide more implementation details and more information about the dataset.
In Section \ref{b}, we provide two more instantiations of our model, i.e., the MIA-VSR-small and MIA-VSR-tiny to compare with other light-weight VSR models.
In Section \ref{c}, we provide a further result of fine-tuning our MIA-VSR model by longer sequences. 
More visual examples of different VSR models are presented in Section \ref{d}.

\section{Dataset and implementation details}\label{a}
\subsection{Datasets}
\textbf{REDS \cite{nah2019ntire}}
REDS is a widely-used video dataset for evaluating video restoration tasks.
It has 270 clips with a spatial resolution of 1280 × 720. 
We follow the experimental settings of \cite{chan2020basicvsr,chan2022basicvsr++,shi2022rethinking} and use REDS4 (4 selected representative
clips, i.e., 000, 011, 015 and 020) for testing and training our models on the remaining 266 sequences.

\vspace{3mm}
\noindent
\textbf{Vimeo-90K \cite{xue2019video}} Vimeo-90K is a commonly used dataset which contains 64,612 training clips and 7,824 testing clips (denoted as Vimeo90K-T).
Each clip contains 7 frames of images with a spatial resolution of 448 $\times$ 256.
We follow the experimental settings of \cite{chan2020basicvsr,chan2022basicvsr++,shi2022rethinking} and evaluate our proposed MIA-VSR method with the Vimeo-90K dataset.

\vspace{3mm}
\noindent
\textbf{Vid4 \cite{liu2013bayesian}} Vid4 is a classical dataset for evaluating video super-resolution methods.
It contains 4 video clips (i.e., calendar,
city, foliage and walk) and each clip has at least 34 frames (720 × 480).
We follow the experimental settings of \cite{chan2020basicvsr,chan2022basicvsr++,shi2022rethinking} and use the 4 sequences in the Vid4 dataset to compare different VSR models.

\subsection{{Training and testing details}}
\noindent
\textbf{Implementation details for the REDS model.} 
We train our MIA-VSR model with the REDS \cite{nah2019ntire} training
dataset with zooming factor 4.
We follow the experimental settings of BasicVSR++ \cite{chan2022basicvsr++} and train our MIA-VSR model for 600K iterations.
The initial learning rate is set as $2 \times 10^{-4}$. 
We train our model with Adam optimizer and the batch size is set as 24.
In the testing phase, we evaluate MIA-VSR model's performance on the REDS4 \cite{nah2019ntire} dataset.

\vspace{3mm}
\noindent
\textbf{Implementation details for the Vimeo-90K and Vid4 model.}
We train our MIA-VSR model with the Vimeo90K \cite{nah2019ntire} training
dataset with zooming factor 4.
We follow the experimental settings of BasicVSR++ \cite{chan2022basicvsr++} and train our MIA-VSR model for another 300K iterations with its well-trained model on the REDS dataset.
The initial learning rate is set as $1 \times 10^{-4}$. 
We train our model with Adam optimizer and set the batch size at
24.
In the testing phase, we evaluate the performance of the MIA-VSR model on Vimeo90K-T \cite{xue2019video} and Vid4 \cite{liu2013bayesian} datasets.

\section{Light-weight MIA-VSR models }\label{b}
In order to compare with recently proposed light-weight methods,
we establish two light-weight versions of our MIA-VSR model, i.e., the MIA-VSR-small and the MIA-VSR-tiny model.
Both the MIA-VSR-small and the MIA-VSR-tiny models contain 4 feature propagation modules and each feature propagation module comprises 6 MIIA blocks with a skip connection.
The spatial window size and the head size are set to $8\times 8$ and 6.
While, the major difference between the two models lies in their respective numbers of channels, we set the channel number of MIA-VSR-small as 120 and set the channel number of
MIA-VSR-tiny as 96.

\begin{figure*}[h]
\centering
\includegraphics[width=0.68\textwidth]{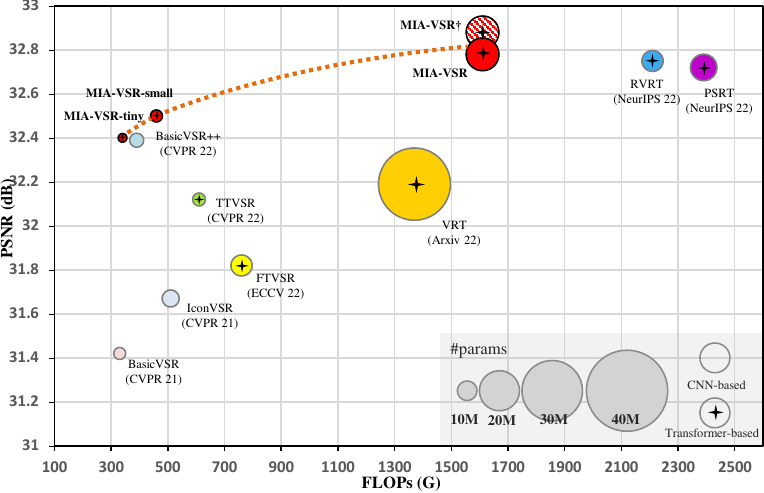}
\caption{\textbf{PSNR(dB) and FLOPs(G) comparison on the REDS4 \cite{nah2019ntire} dataset.} 
In comparison with the existing video super-resolution methods, our proposed MIA-VSR model, MIA-VSR-small and MIA-VSR-tiny could obtain better trade-offs between VSR results and computational cost.
Our fine-tuned model MIA-VSR$\dagger$ outperforms the current state-of-the-art model by more than 0.1 dB with nearly 40\% less number of FLOPs.  
Our light-weight model MIA-VSR-tiny outperforms the recent light-weight Transformer-based VSR model TTVSR\cite{liu2022learning} by 0.28 dB, with 45\% less number of FLOPs.
More details can be found in Section \ref{b}.
}
\label{fig:redscompare}
\end{figure*}

The PSNR values, number of parameters and running FLOPs by different VSR models are presented in Fig.\ref{fig:redscompare} and Table \ref{tab:moresota-table}.
In addition to our comparison models in the main paper, two recently proposed efficient VSR models, i.e., FTVSR \cite{qiu2022learning} and TTVSR \cite{liu2022learning}, are
also included for reference.
Generally, Transformer-based VSR models could achieve better VSR results than CNN-based methods.
MIA-VSR outperforms the state-of-the-art 
CNN-based model BasicVSR++ by a large margin.
Furthermore, with less number of parameters and less running FLOPs, our MIA-VSR-tiny model could achieve a better trade-off between computation burden and VSR results over the existing CNN-based models.
In comparison with state-of-the-art Transformer-based methods, 
our MIA-VSR model could achieve better VSR results with much less computational cost.
While, our light-weight models MIA-VSR-small and MIA-VSR-tiny also achieved a better trade-off between VSR results and computational cost
than the existing light-weight Transformer-based methods;
with 45\% less number of FLOPs, our MIA-VSR-tiny could improve the TTVSR model by 0.28 dB.
\begin{table*}[h]
	\renewcommand{\arraystretch}{0.8}
	\renewcommand{\tabcolsep}{3.2pt}
    \caption{Quantitative comparison (PSNR/SSIM) on the REDS4 \cite{nah2019ntire}, Vimeo90K-T \cite{xue2019video} and Vid4 \cite{liu2013bayesian} dataset for 4× video super-resolution task. For each group of experiments, we color the best and second-best performance with \textcolor{red}{red} and \textcolor{blue}{blue}, respectively.}
    \label{tab:moresota-table}
      \begin{threeparttable}
    \begin{adjustbox}{center}
    \centering
    \begin{tabular}{c|c|c||ccc|ccc|ccc}
    \toprule
    \multirow{2}{*}{Method} & Frames&Params  & \multicolumn{3}{c}{REDS4} & \multicolumn{3}{c}{Vimeo-90K-T} & \multicolumn{3}{c}{Vid4}  \\
    & REDS/Vimeo&(M)& PSNR & SSIM&FLOPs & PSNR & SSIM&FLOPs & PSNR & SSIM&FLOPs    \\
    \midrule
    BasicVSR \cite{chan2020basicvsr}     & 15/14& 6.3 & 31.42 & 0.8909& 0.33& 37.18 & 0.9450& 0.041 & 27.24 & 0.8251&0.134\\
    IconVSR \cite{chan2020basicvsr}      & 15/14& 8.7  & 31.67 & 0.8948& 0.51& 37.47 & 0.9476& 0.063 & 27.39 & 0.8279&0.207\\
    BasicVSR++ \cite{chan2022basicvsr++} & 30/14&7.9  & 32.39 & 0.9069& 0.39& 37.79 & 0.9500& 0.049 & 27.79 & 0.8400&0.158 \\
    \midrule
    \midrule
    FTVSR \cite{qiu2022learning} & 40/- & 10.8 & 31.82& 0.8960 & 0.76&-&-&-&-&-&- \\
    TTVSR \cite{liu2022learning}&50/-& 6.8 &32.12&0.9021&0.61&-&-&-&-&-&-\\
    \rowcolor{gray!20}
    MIA-VSR-tiny & 80/-& 4.8 & \textcolor{blue}{32.40} & \textcolor{blue}{0.9176} & 0.35& - & -& -& -& -& -  \\
    \rowcolor{gray!20}
    MIA-VSR-small & 50/-& 6.3 & \textcolor{red}{32.50} & \textcolor{red}{0.9197} & 0.47 & - & -& -& -& -& -  \\
    \midrule
    VRT \cite{liang2022vrt}  & 16/7& 35.6 & 32.19 & 0.9006& 1.37& 38.20 & 0.9530& 0.170 & 27.93 & 0.8425&0.556\\
    RVRT \cite{liang2022recurrent}       & 30/14& 10.8 & 32.75 & 0.9113& 2.21& 38.15 & 0.9527& 0.275 & 27.99 & 0.8462&0.913 \\
    PSRT-recurrent \cite{shi2022rethinking} & 16/14& 13.4 & 32.72 & 0.9106&2.39 &\textcolor{red}{38.27} & \textcolor{red}{0.9536}&0.297&\textcolor{blue}{28.07}& \textcolor{blue}{0.8485}&0.970\\
    \rowcolor{gray!20}
    MIA-VSR& 16/14& 16.5 & \textcolor{blue}{32.78} & \textcolor{blue}{0.9220}&1.61 & \textcolor{blue}{38.22}& \textcolor{blue}{0.9532} &0.204& \textcolor{red}{28.20} & \textcolor{red}{0.8507}&0.624\\
    \rowcolor{gray!20}
    MIA-VSR$\dagger$& 40/-& 16.5 & \textcolor{red}{32.88} & \textcolor{red}{0.9241}&1.61 & -&-&-& -&-&-\\
    \bottomrule
    \end{tabular}
  \end{adjustbox}
    \begin{tablenotes} 
	\footnotesize \item MIA-VSR$\dagger$ is fine-tuned with the well-trained MIA-VSR model by 40 frames from the REDS \cite{nah2019ntire} dataset.
    \end{tablenotes}
    \end{threeparttable}
\end{table*}
\section{Fine-tune MIA-VSR with longer sequences.} \label{c}
For further proving that training the VSR model with longer sequences can get a better result, we chose the MIA-VSR model which is trained for 450K iterations with 16 frames from the REDS \cite{nah2019ntire} dataset and fine-tuned it for another 150K iterations with 40 frames, named MIA-VSR$\dagger$. The comparison with the state-of-the-art Transformer-based VSR methods (i.e., VRT \cite{liang2022vrt}, RVRT \cite{liang2022recurrent} and PSRT \cite{shi2022rethinking}) can be found in Fig.\ref{fig:redscompare} and Table \ref{tab:moresota-table}. It has a further improvement of the MIA-VSR model trained with 16 frames from the REDS dataset by 0.1dB without adding another the number of FLOPs. 
\section{Visual results}\label{d}
\label{morevisual}
\begin{figure*}
\centering
\includegraphics[width=0.9\textwidth]{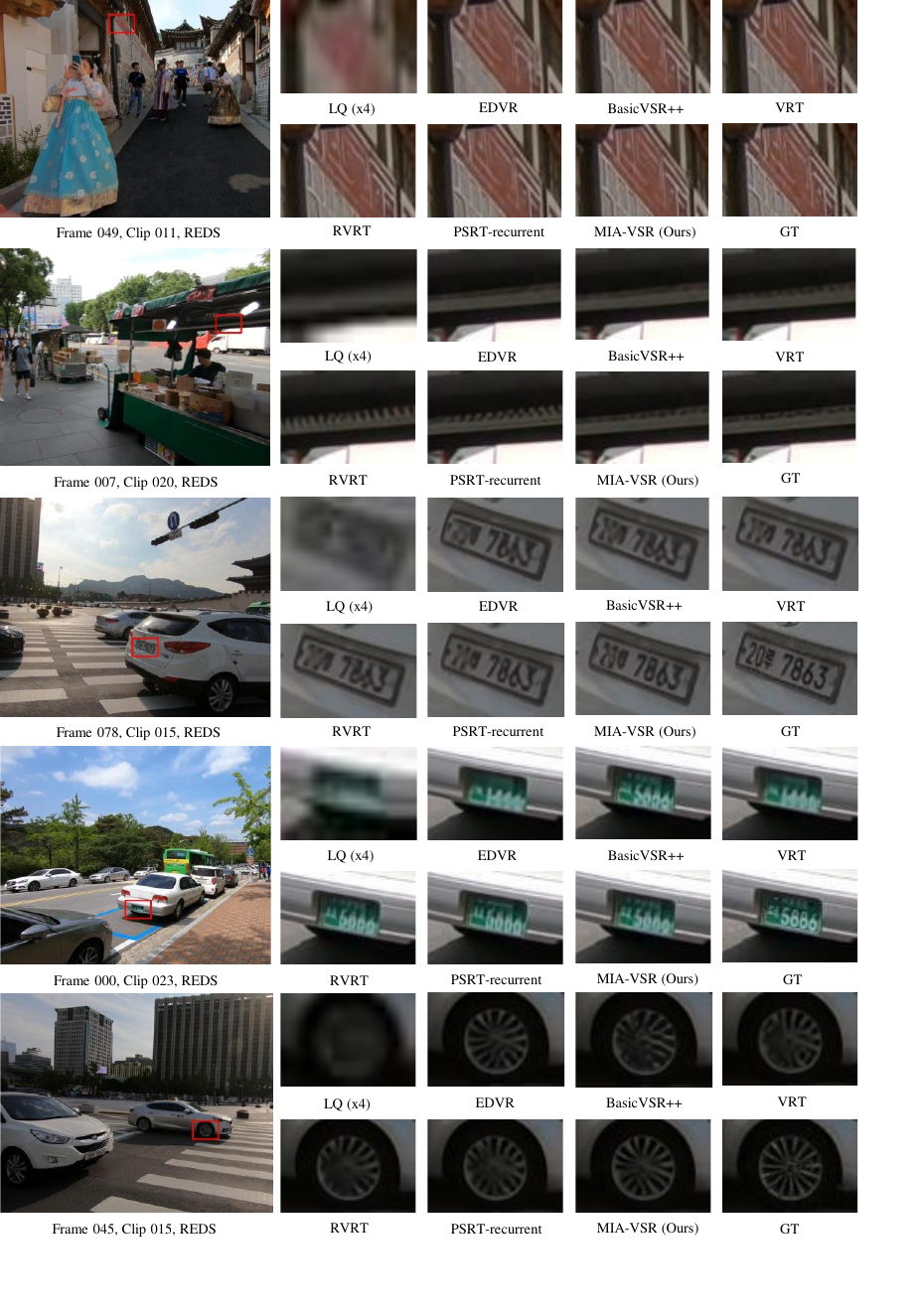}
\caption{Visual comparison for $4\times$ VSR on REDS4 dataset.}
\label{fig:morevisualreds}
\end{figure*}

\begin{figure*}
\centering
\includegraphics[width=0.9\textwidth]{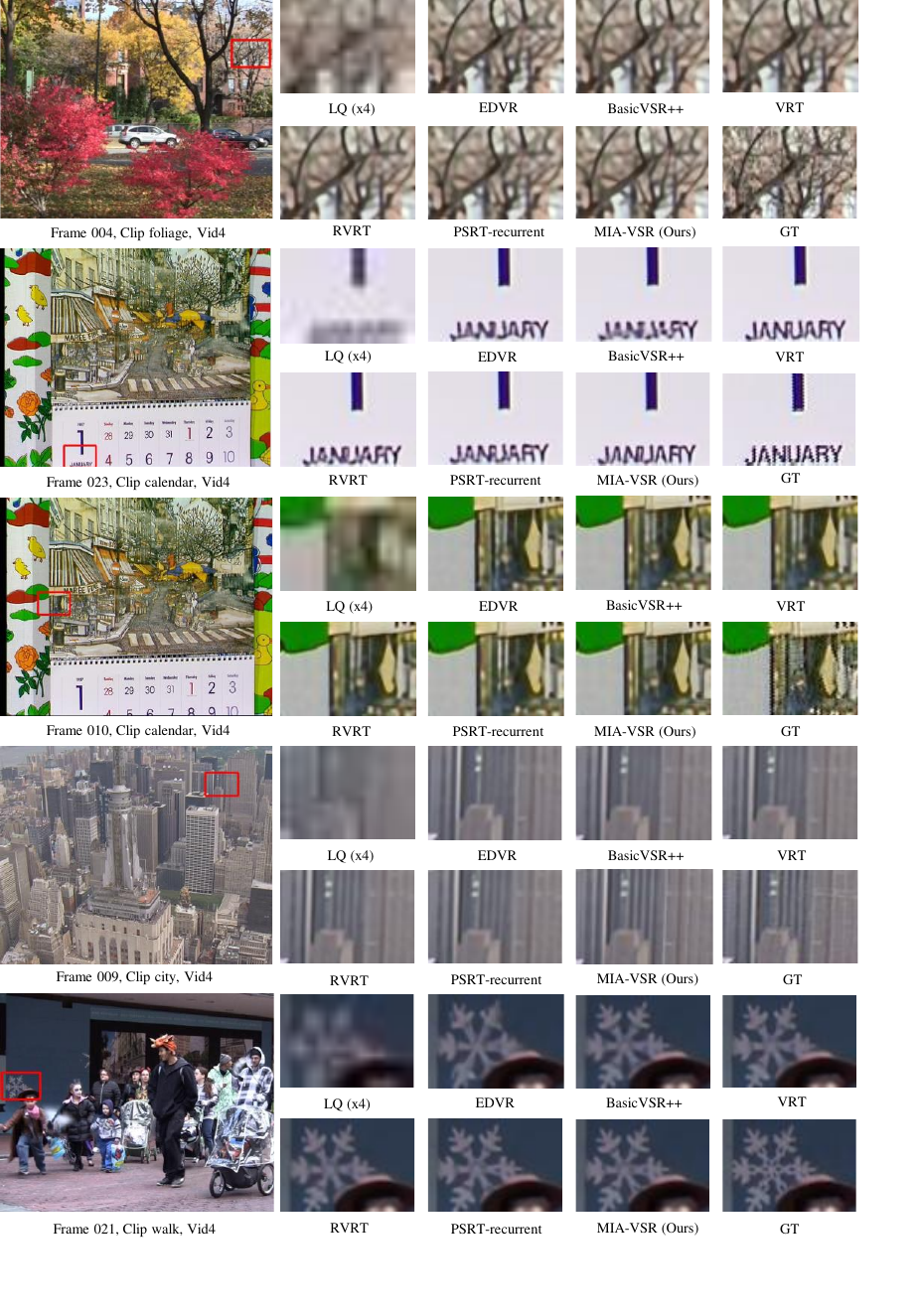}
\caption{Visual comparison for $4\times$ VSR on Vid4 dataset.}
\label{fig:morevisualvid4}
\end{figure*}

We show more visual comparisons between the existing VSR methods and the proposed VSR Transformer with masked inter\&intra-frame attention (MIA). We use 16 frames to train on the REDS dataset
and 14 on the Vimeo-90K dataset. Fig.\ref{fig:morevisualreds} and Fig.\ref{fig:morevisualvid4} show the visual results. It can be seen that, in addition
to its quantization improvement, the proposed method can generate visually pleasing images with
sharp edges and fine details, such as horizontal bar patterns of buildings and numbers on license
plates. On the contrary, existing methods suffer from texture distortion or loss of detail in these scenes.
\end{document}

%% file: preamble.tex
\usepackage[dvipsnames]{xcolor}
